\documentclass{article}

\PassOptionsToPackage{numbers, compress}{natbib}

\usepackage[preprint]{neurips_2026}

\usepackage[utf8]{inputenc} 
\usepackage[T1]{fontenc}    
\usepackage{CJKutf8}        
\usepackage{hyperref}       
\usepackage{url}            
\usepackage{booktabs}       
\usepackage{amsfonts}       
\usepackage{nicefrac}       
\usepackage{microtype}      
\usepackage{xcolor}         
\usepackage{textcomp}
\usepackage{graphicx}
\usepackage{multirow}
\usepackage{color}
\usepackage{threeparttable} 
\usepackage{tablefootnote} 
\usepackage{arydshln} 
\usepackage{amssymb} 
\usepackage{pifont} 
\usepackage{wrapfig}
\usepackage{verbatim}
\usepackage{enumitem}
\usepackage{amsmath}
\usepackage{colortbl}
\usepackage{setspace}
\usepackage{hhline}
\usepackage{subcaption}
\usepackage{makecell}
\usepackage{mathrsfs}
\usepackage{tikz}
\usepackage{tabularx}
\usepackage{fontawesome5}
\usepackage{algorithm}
\usepackage{algorithmic}

\newcommand{\redx}{{\color{red}\ding{55}}} 

\newcommand{\MYBENCH}{\textsc{QuestBench}}  

\newcommand{\redtext}[1]{#1}

\newcolumntype{C}[1]{>{\centering\arraybackslash}p{#1}}
\title{Teaching AI Through Benchmark Construction: \MYBENCH~as a Course-Based Practice for Accountable Knowledge Work}

\author{%
  \vspace{-25pt}\\
  \textbf{Haiyang Shen$^{1,*,\dag}$\quad Jiuzheng Wang$^{1,*,\dag}$\quad Taian Guo$^{1}$\quad Mugeng Liu$^{1}$}\\
  \textbf{Wenchun Jing$^{1}$\quad Chongyang Pan$^{1}$\quad Siqi Zhong$^{1}$\quad Zhiyang Chen$^{1}$}\\
  \textbf{Weichen Bi$^{1}$\quad Yudong Han$^{1}$\quad Xiaoying Bai$^{2}$\quad Yun Ma$^{1,\dag}$}\vspace{8pt}\\
  $^{1}$Peking University \quad $^{2}$Advanced Institute of Big Data \\
  \texttt{\small hyshen@stu.pku.edu.cn} \quad \texttt{\small jiuzhengwang@stu.pku.edu.cn} \quad \texttt{\small mayun@pku.edu.cn}\\
  \vspace{3em}
  $^*$Equal contribution, \textsuperscript{\dag}Corresponding Authors
  \vspace{-48pt} \\
}

\begin{document}

\maketitle

\begin{abstract}

As AI becomes part of everyday learning, many courses teach students to use it mainly as a productivity tool: how to prompt, search, summarize, write, code, and use tools more efficiently. We argue that AI education also needs a setting in which students learn to test AI and understand their own role in judging machine-produced knowledge. To this end, we introduce a course-based practice that teaches AI through benchmark construction, using deep research systems as a concrete example of AI-era knowledge work. Students turn disciplinary knowledge into verifiable expert-level questions, review one another's designs for ambiguity and shortcuts, and evaluate AI systems on the resulting tasks. This activity gives students direct exposure to a powerful tool while asking them to specify what a trustworthy answer would require. The produced benchmark, \MYBENCH, consists of 256 questions across 14 humanities and social-science domains. Evaluation on \MYBENCH~shows that student-designed tasks reveal hidden failures in current deep research systems: across \redtext{thirteen} evaluated systems, the mean question-level pass rate is only \redtext{16.85\%}, and the best-performing system, GPT-5.5, reaches a \redtext{57.58\%} pass rate. The failures are educationally useful because they show how fluent, source-backed answers can still miss the right query, source, term, or evidence standard. Reflections from five student contributors suggest that benchmark construction can help students see professional knowledge not only as content AI may retrieve, but as the basis for judging AI outputs. We present \MYBENCH~as a benchmark artifact and as a reusable classroom setting for a larger educational question: how students can remain responsible knowledge actors as AI enters learning and professional work. The dataset is available at \url{https://huggingface.co/datasets/PKUAIWeb/QuestBench/tree/main}.

\end{abstract}

\section{Introduction}
\label{sec:Introduction}

AI is becoming part of how students do knowledge work. Current systems can search, read, write, code, call tools, and help complete tasks that once required many separate human steps. This shift does not remove students from the work. It changes what they must be able to do within it. Much AI instruction begins with tool use: how to prompt a system, obtain useful output, and work more efficiently. That first step is necessary, but it is incomplete. If students only learn to receive AI outputs, they may not learn how to define the task, inspect the process, check the evidence, or decide whether the result can be trusted. A central challenge for AI education is therefore to teach students how to remain accountable when AI participates in producing knowledge.

We propose benchmark construction as a course-based way to teach this responsibility. Building a benchmark does not ask students to stand outside AI work as passive evaluators. It asks them to build the conditions under which AI work can be judged. Students must decide what is worth asking, what counts as a valid answer, which sources can support it, which shortcuts would make the task meaningless, and which field-specific distinctions the grading criteria must preserve. These decisions are part of responsible AI-mediated work in professional settings. Benchmark construction makes them explicit enough to practice in a classroom.

This paper uses deep research systems as the teaching case. They are not the endpoint of the argument; they are a concrete example of AI-mediated knowledge work. Such systems search the web, visit documents, synthesize evidence, and return answers that often appear well grounded. They are useful enough for students to take seriously and unreliable enough to teach from. Their fluent and source-backed outputs can hide failures that only domain-aware users can recognize, such as a wrong query, a shortcut, a confused source, an outdated term, or partial evidence presented as a complete answer.

\begin{figure}[t]
    \centering
    \includegraphics[width=\linewidth]{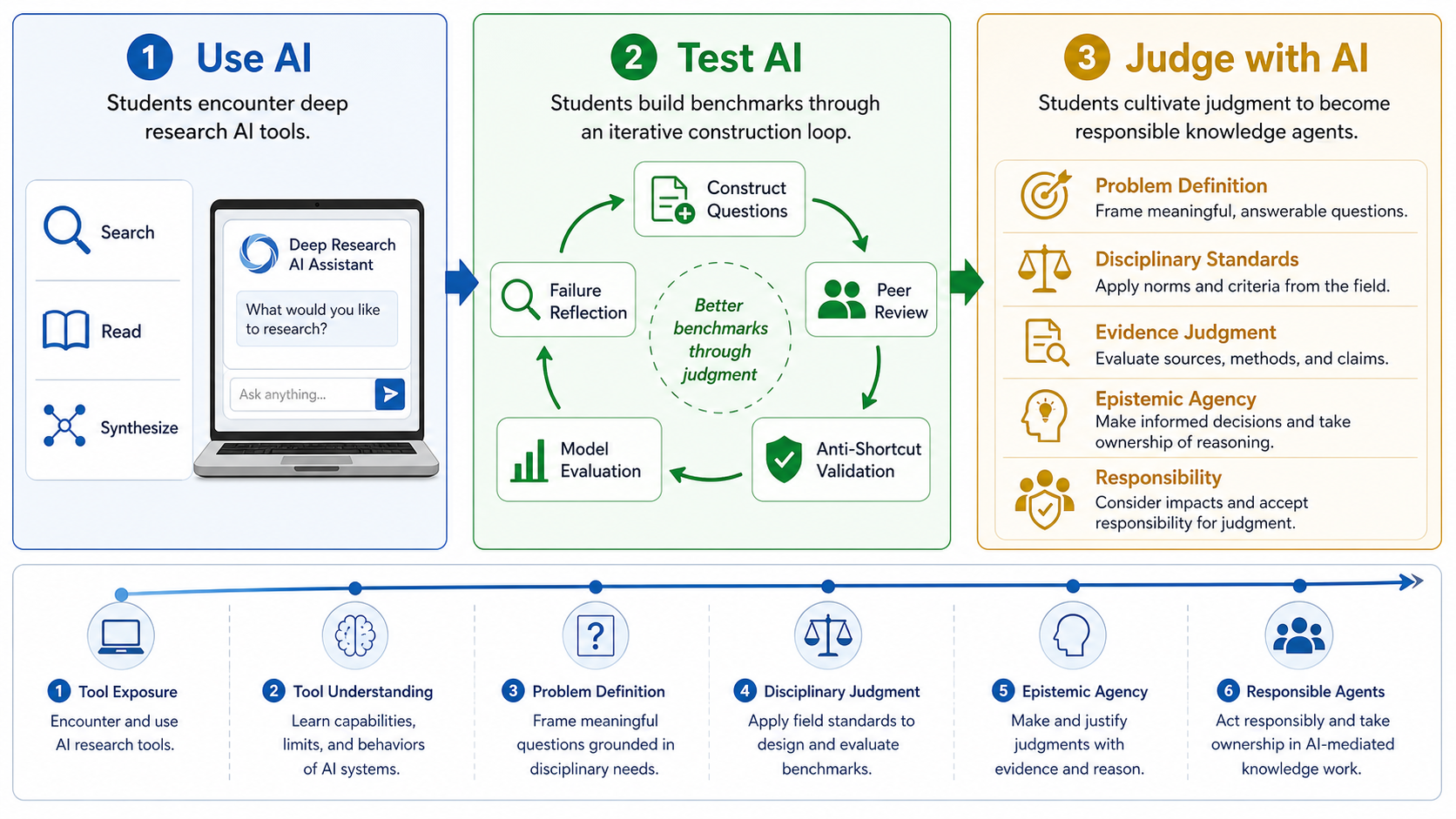}
    \caption{Conceptual framework of \MYBENCH~as course-based benchmark construction for teaching accountable AI-mediated knowledge work. Students first encounter deep research systems as a practical tool, then use benchmark construction to design expert-level questions, test shortcuts, validate answers, evaluate models, and analyze failures. The course links tool exposure with question design, disciplinary standards, and responsibility for judging AI-produced work.}
    \label{fig:framework}
\end{figure}

This approach gives disciplinary knowledge a specific role in AI education. A narrow view treats such knowledge as content that AI may retrieve or summarize. In \MYBENCH, students use it to define the standards under which AI-produced work can be accepted. Law students know why statutory wording and versioning matter. History and international-relations students know why provenance and document identity matter. Language and literature students know why translation, editions, and phrasing matter. These details are operational, but they also carry the standards through which information becomes trustworthy knowledge.

The educational aim is to help students remain responsible knowledge actors when AI participates in the work. AI systems can search and synthesize, but they do not decide which questions deserve attention, what evidence a field should accept, or when an answer is good enough to rely on. As AI becomes more capable, education has to teach both use and accountability. The goal is not to replace making with judging. Rather, students need to learn how question design, evidence selection, model use, and answer acceptance form one chain of responsibility. Benchmark construction makes that chain concrete by asking students to define, test, and defend the standards by which AI outputs will be judged.

We instantiate this approach in \MYBENCH, a course-based benchmark construction project for teaching accountable use of deep research systems. Students from humanities and social-science disciplines at Peking University designed, reviewed, and validated challenging information-seeking questions from their own fields. The final benchmark contains 256 curated questions spanning 14 normalized domains, including law, history, international relations, literature, social sciences, arts, and foreign languages. Each question is publicly answerable but difficult for current systems because success requires domain-aware query formulation, specialized source navigation, evidential judgment, and precise answer extraction. The result is both a benchmark and a record of a course practice: it evaluates deep research systems while showing how students can learn to define and inspect AI-mediated knowledge work.

Evaluation on \MYBENCH~shows that student-designed tasks reveal clear limitations in current deep research systems. We evaluate Kimi K2.5~\cite{kimiteam2026k25} and DeepSeek-V3.2~\cite{deepseekai2025v32}, alongside Seed-2 Pro~\cite{bytedanceseed2026seed20} and Seed-1.8 Pro~\cite{bytedanceseed2026seed18}\redtext{, together with nine frontier deep-search systems: GPT-5.5~\cite{openai2026gpt55}, Claude Opus 4.7~\cite{anthropic2026opus47}, Gemini 3.1 Pro~\cite{google2026gemini31}, GLM 5.1~\cite{zhipu2026glm51}, DeepSeek-V4 Pro~\cite{deepseekai2026v4pro}, Kimi K2.6~\cite{kimiteam2026k26}, MiMo-V2.5 Pro~\cite{xiaomi2026mimo25}, Qwen 3.6 Plus~\cite{qwen2026qwen36}, and MiniMax M2.7~\cite{minimax2026m27}}. Mean scores range from \redtext{14.58 to 67.12} out of 100, and pass rates range from \redtext{7.81\% to 57.58\%}. Failure analysis shows that the main bottleneck is not information availability alone, but the interaction of query formulation, source navigation, and answer extraction under discipline-specific standards. For the course, these failures are more than model errors. They are cases through which students can inspect where the AI-mediated work broke down and what human responsibility remains.

The main contributions are:
\begin{itemize}[leftmargin=*,nosep]
    \item We introduce benchmark construction as a course-based approach to teaching accountable AI-mediated knowledge work. The activity begins with direct exposure to an AI-era productivity tool, then asks students to design verifiable tasks, defend evidence, and inspect model failures.
    \item We present \MYBENCH, a benchmark and course artifact containing 256 expert-level deep research questions across 14 humanities and social-science domains. The construction process combines student disciplinary expertise, adversarial peer review, anti-shortcut validation, and multi-round quality control.
    \item We evaluate \redtext{thirteen} state-of-the-art deep search systems on \MYBENCH~and identify recurring failure patterns including retrieval failure, unsupported inference, entity confusion, and answer-extraction errors. These results show that student-designed tasks can make hidden AI failures observable.
    \item We use reflections from five student contributors to discuss how students understand tools, define problems, judge evidence, and locate their responsibility in AI-mediated knowledge work. We treat this as part of a continuing inquiry into AI's educational impact.
\end{itemize}

\section{Background and Positioning}
\label{sec:RelatedWork}

This paper treats benchmark construction as both an evaluation practice and a teaching practice for accountable AI-mediated knowledge work. We therefore organize related work around three questions: what kind of AI work students need to learn to inspect, why such work requires expert evaluation, and how benchmark construction can become part of AI education.

\subsection{AI Education Beyond Tool Use}
\label{subsec:ai_education}

AI education often begins with use. Students learn how to prompt systems, obtain useful outputs, and apply AI to reading, writing, programming, or research tasks. This first contact matters because students need to work with tools that are entering knowledge work. Tool use alone, however, does not teach students how to define a task, inspect the process that produced an answer, check whether the evidence is sufficient, or decide whether the result can be used responsibly.

\MYBENCH~addresses this gap by treating AI as both a tool and an object of study. The course begins with a concrete productivity tool, then asks students to define evaluation tasks, specify answer standards, test shortcuts, and analyze failures. Benchmark construction is therefore the teaching activity rather than only the data collection procedure. It turns accountability from an abstract norm into a set of operations students can practice.

\subsection{Deep Research as a Concrete AI Teaching Case}
\label{subsec:deep_search}

Deep research systems are a useful case for this educational question because they perform several parts of knowledge work in one workflow. Earlier benchmarks such as Natural Questions~\cite{kwiatkowski2019natural}, TriviaQA~\cite{joshi2017triviaqa}, and HotpotQA~\cite{yang2018hotpotqa} evaluate retrieval-grounded and multi-hop answering. Search-augmented evaluations such as FreshLLMs~\cite{vu2023freshllms} and GAIA~\cite{mialon2024gaia} move toward dynamic tool-using assistants. More recent work studies agentic search: BrowseComp~\cite{wei2025browsecomp,zhou2025browsecompzh} and DeepSearchQA~\cite{gupta2026deepsearchqa} assess search-intensive QA, while ResearchArena~\cite{kang2025researcharena} and DeepResearch Bench~\cite{du2025deepresearchbench} evaluate research agents. RAG-focused work such as FRAMES~\cite{krishna2025frames} and RAGChecker~\cite{ru2024ragchecker} provides finer-grained retrieval diagnosis.

These systems are no longer simple answer engines. They search, read, synthesize, cite, and decide when a response appears complete. That makes them powerful productivity tools, and it also makes their failures harder for students to notice. A source-backed answer may look like knowledge even when it rests on a wrong query, a confused document, a missing source, or an imprecise term. \MYBENCH~uses deep research as a representative case for teaching students how to inspect the AI-mediated work process behind a final answer.

\subsection{Expert Evaluation as a Source of Educational Standards}
\label{subsec:expert_knowledge}

Expert-level benchmarks show why general tool use is not enough. GPQA~\cite{rein2024gpqa} targets graduate-level questions validated through an expert-vs-non-expert gap. Humanity's Last Exam~\cite{phan2026hle} curates questions intended to be unsolvable by current AI. MMLU-Pro~\cite{wang2024mmlupro} and SuperGPQA~\cite{du2025supergpqa} scale closed-ended academic difficulty. PaperBench~\cite{starace2025paperbench} tests research artifact replication, while AgentBench~\cite{liu2024agentbench} and $\tau$-bench~\cite{yao2025taubench} broaden tool-use evaluation. Domain-specific benchmarks such as LegalBench~\cite{guha2023legalbench}, PubMedQA~\cite{jin2019pubmedqa}, and FinQA~\cite{chen2021finqa} evaluate specialized reasoning within individual fields, while WebArena~\cite{zhou2024webarena} and Mind2Web~\cite{deng2023mind2web} study task completion in grounded web environments.

\MYBENCH~draws from this evaluation tradition, but uses expert-level difficulty for a course purpose. Students have to specify what makes a question meaningful, what makes an answer verifiable, what evidence counts, and what distinctions cannot be collapsed. The same properties that make a question hard for AI systems, including specialized sources, precise terminology, anti-shortcut structure, and explicit grading criteria, also make it useful for teaching students how to hold AI-mediated knowledge work accountable.

\subsection{Benchmark Construction as AI Pedagogy}
\label{subsec:benchmark_pedagogy}

Most benchmark work treats construction as a research methodology: experts or annotators create tasks so that models can be evaluated. \MYBENCH~keeps this research function and adds a course function. Students build the conditions under which AI outputs can be judged: they define tasks, defend answers, audit shortcuts, and score model outputs under explicit standards.

This framing differs from tool-oriented AI literacy. Prompting and efficient use matter, but they do not by themselves teach students how knowledge becomes trustworthy. Benchmark construction asks students to specify the conditions under which an AI answer should count as correct. That is where the educational value sits: problem definition, evidence discipline, peer scrutiny, and responsibility for judgment become part of the assignment.

Table~\ref{tab:benchmark_comparison} summarizes high-level benchmark properties. We include it to show both the evaluation setting and the research substrate that the course adapts for teaching.

\begin{table}[t]
\centering
\small
\caption{Comparison of \MYBENCH~with related benchmarks across key dimensions. H\&SS = Humanities \& Social Sciences.}
\label{tab:benchmark_comparison}
\begin{tabular}{lcccccc}
\toprule
Benchmark & Size & Domains & Web Search & Anti-Shortcut & Authors & Validation \\
\midrule
BrowseComp & 1,266 & General & \checkmark & \checkmark & Researcher & Internal \\
DeepSearchQA & 900 & 17 fields & \checkmark & \redx & Researcher & Internal \\
GAIA & 466 & General & \checkmark & \redx & Researcher & Internal \\
GPQA & 448 & STEM & \redx & \checkmark & PhD student & Expert gap \\
HLE & 3,000 & Broad & \redx & \redx & Expert & Peer review \\
\midrule
\MYBENCH & 256 & 14 (H\&SS) & \checkmark & \checkmark & Univ. students & Multi-round \\
\bottomrule
\end{tabular}
\end{table}

The comparison clarifies the dual nature of \MYBENCH. As a benchmark, it evaluates open-web deep research across expert-level professional domains. As a course practice, it asks students to learn AI by defining, testing, and judging the tasks that expose where AI-mediated work can fail.

\section{Course Design: Teaching AI Through Benchmark Construction}
\label{sec:Method}

\MYBENCH~is designed as a course practice first and a benchmark artifact second. The course asks a practical educational question: if students are already learning and working with AI tools, how can they remain responsible for the knowledge work those tools help produce? We make benchmark construction the main learning activity. Students encounter deep research as a concrete AI productivity tool, transform disciplinary knowledge into verifiable evaluation tasks, review one another's tasks for ambiguity and shortcuts, and use model failures to discuss where AI-mediated work can and cannot be trusted.

\subsection{Pedagogical Design}
\label{subsec:pedagogical_design}

The course is organized around a progression from exposure to accountable inspection. Students first encounter AI as a productivity tool. Deep research systems provide the concrete case because they search, read, synthesize, and answer in ways that resemble future knowledge work. Students then study the tool by designing questions that it must answer under explicit standards. This moves them from asking AI for answers to specifying a valuable, bounded, and verifiable task. Finally, students analyze failures and ask which part of the work required disciplinary knowledge: the query, the source, the term, the answer boundary, or the decision to keep searching. The progression keeps the first layer of AI education, direct contact with the tool, while making room for the larger question of how students remain responsible knowledge actors when AI participates in their work.

Benchmark construction fits this design because it asks students to build the standards by which AI work can be inspected. Instead of receiving answers from a system, students decide what would count as a rigorous test of the system. They must externalize tacit disciplinary knowledge: what sources their field trusts, which distinctions matter, what would make an answer incomplete, and where a model might exploit a shortcut. The course treats this work as a bridge between learning to use AI and learning to remain accountable for AI-mediated knowledge work.

\subsection{Deep Research as the Teaching Case}
\label{subsec:deep_research_case}

The course could use many AI tools as teaching objects. We use deep research because it is a consequential example of AI-era knowledge work. These systems do more than generate text; they search, visit documents, synthesize evidence, and present answers that often appear well grounded. This makes them useful for students and, at the same time, educationally risky. A fluent answer with citations may encourage students to treat output as knowledge before asking whether the question was well posed, whether the sources were authoritative, or whether the answer satisfies the standards of a field.

Deep research therefore functions as a case for studying a larger issue: how AI tools should be situated within human standards of inquiry. When students build questions that expose hidden failures, they also learn why those failures matter. An incorrect legal version, a confused archival document, a missing source, or an imprecise translation is a model error with professional consequences. It matters because a discipline has standards for what counts as knowledge, and those standards remain part of the student's responsibility when AI helps produce an answer.

\subsection{Course-Based Construction Activity}
\label{subsec:course_construction}

\begin{figure}[t]
    \centering
    \includegraphics[width=\linewidth]{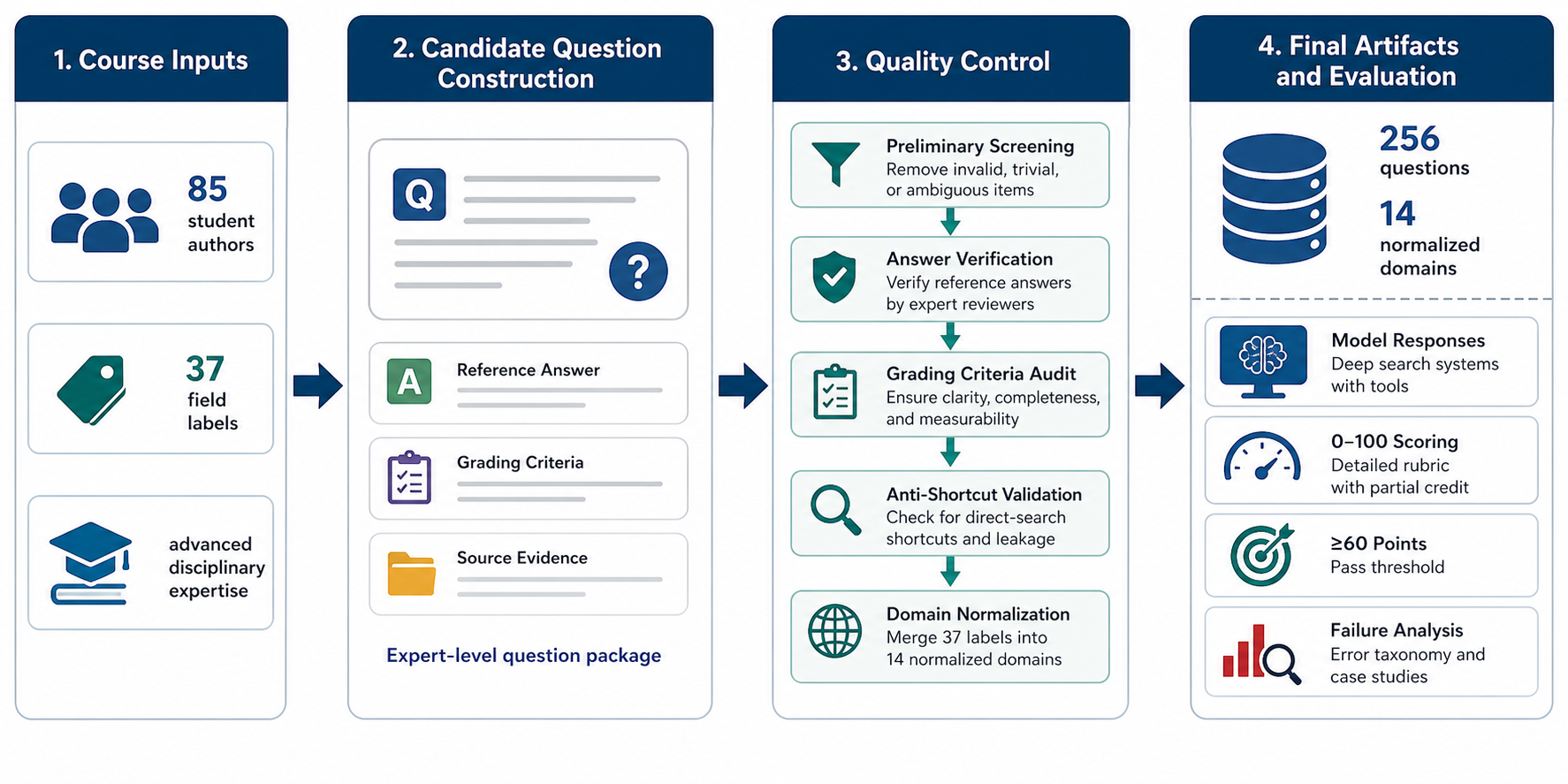}
    \caption{Course and technical pipeline for \MYBENCH. Students transform disciplinary knowledge into expert-level question packages, then filter them through preliminary screening, answer verification, grading-criteria audit, anti-shortcut validation, and domain normalization. The same artifacts are then used for model evaluation, scoring, and failure analysis, turning task design into practice in accountable AI-mediated knowledge work.}
    \label{fig:pipeline}
\end{figure}

We collaborate with course instructors at Peking University to recruit undergraduate students across academic disciplines. \MYBENCH~contains questions from 85 student question authors, covering 37 self-reported field labels that we normalize into 14 domain groups, including law, history, international relations, literature, foreign languages, social sciences, journalism, and archaeology. All participants have completed advanced coursework in their respective fields. This disciplinary spread matters for the course design because each field gives students a different way to examine how AI tools depend on human standards of inquiry.

The course project asks students to construct questions from their own domains rather than generic trivia questions. This choice improves benchmark quality because students know the sources, terminology, and reasoning patterns of their fields. It also gives the course its educational force. Students must articulate how their training determines standards for reliable answers in AI-mediated work. A law question may require exact statutory wording and version awareness; a history question may require archival provenance; a literature or language question may require attention to editions, translation, and phrasing. These requirements show that professional knowledge is more than content for AI to retrieve. It is a structure for deciding whether an AI-produced result can be accepted.

\subsection{Question Design as an Educational Protocol}
\label{subsubsec:question_design}

Without systematic guidelines, crowdsourced questions often have predictable problems: answers available through simple searches, ambiguous grading criteria, or artificial complexity that does not reflect genuine difficulty. We therefore require question designers to satisfy five requirements: (1)~\textit{domain expertise}, so that questions require specialized knowledge inaccessible to educated non-experts; (2)~\textit{long-tail information targeting}, so that answers are embedded in specialized sources rather than obtainable through general knowledge; (3)~\textit{answer uniqueness and verifiability}, with unambiguous grading criteria and documented source materials; (4)~\textit{complexity evolution documentation}, which records how simple questions were refined into expert-level queries; and (5)~\textit{anti-shortcut validation}, which requires each question to pass three tests: no pre-existing solutions searchable online, no trivial identifiers that directly determine answers, and no bypassable reasoning steps. The complete protocol with examples appears in Appendix~\ref{app:design_protocol}.

These requirements are annotation rules, but they also do educational work. A verifiable question ties knowledge claims to evidence. A grading criterion makes answer quality depend on explicit standards rather than fluency. Anti-shortcut design teaches students that apparent competence may come from accidental cues rather than genuine problem solving. Construction documentation makes problem formulation visible as intellectual work. Through this protocol, students learn that responsible AI use begins before a model answers: it begins with how the task and its standards are made.

\subsection{Peer Review and Quality Control}
\label{subsubsec:iterative_filtering}

Peer review is central to both the benchmark and the course. Students first construct candidate questions, then test one another's designs by searching for ambiguities, hidden shortcuts, unverifiable answers, and alternative interpretations. This review structure is modeled on scholarly evaluation: claims become stronger when they survive attempts to falsify or bypass them. For students, the process makes clear that accountability is social as well as individual. A trustworthy answer has to withstand examination by others.

To ensure benchmark quality, we systematically filter questions through a three-stage pipeline combining preliminary screening, iterative expert validation, and domain normalization. Algorithm~\ref{alg:iterative_filtering} presents the complete workflow.

\begin{algorithm}[t]
\caption{Iterative Question Filtering and Quality Control}
\label{alg:iterative_filtering}
\small
\begin{algorithmic}[1]
\STATE \textbf{Input:} Candidate question pool $Q = \{q_1, \ldots, q_n\}$
\STATE \textbf{Output:} Curated benchmark $B$
\STATE
\STATE // \textbf{Stage 1: Preliminary Screening}
\STATE $Q_{\mathrm{screened}} \leftarrow$ RemoveInvalidOrTrivial$(Q)$
\STATE
\STATE // \textbf{Stage 2: Expert Validation (Iterative)}
\FOR{$q \in Q_{\mathrm{screened}}$}
    \STATE $\mathrm{reviewers}_1 \leftarrow$ AssignReviewers$(q, 2)$ \hfill // Round 1: Answer Correctness
    \IF{$\neg$ VerifyAnswerCorrectness$(\mathrm{reviewers}_1, q)$}
        \STATE Remove $q$; \textbf{continue}
    \ENDIF
    \STATE $\mathrm{reviewers}_2 \leftarrow$ AssignReviewers$(q, 2)$ \hfill // Round 2: Criteria Audit
    \IF{$\neg$ VerifyCriteriaClarity$(\mathrm{reviewers}_2, q)$}
        \STATE $q \leftarrow$ ReviseGradingCriteria$(q)$; \textbf{restart} Round 1
    \ENDIF
    \STATE $\mathrm{reviewers}_3 \leftarrow$ AssignReviewers$(q, 2)$ \hfill // Round 3: Anti-Shortcut
    \IF{$\neg$ VerifyAntiShortcut$(\mathrm{reviewers}_3, q)$}
        \STATE Remove $q$
    \ENDIF
\ENDFOR
\STATE
\STATE // \textbf{Stage 3: Domain Normalization and Final Assembly}
\STATE $B \leftarrow$ AssembleBenchmark(NormalizeDomains$(Q_{\mathrm{screened}})$)
\STATE \textbf{return} $B$
\end{algorithmic}
\end{algorithm}

\textbf{Stage 1 (Preliminary screening)} removes invalid, ambiguous, or trivially answerable questions, and applies difficulty-based filtering using a baseline search-enabled model. \textbf{Stage 2 (Expert validation)} subjects each remaining question to three independent human review rounds: answer correctness verification, grading criteria audit, and anti-shortcut testing. Questions failing any round are revised or permanently removed. \textbf{Stage 3 (Domain normalization)} merges 37 original field labels into 14 domain groups while preserving meaningful disciplinary distinctions. The final benchmark comprises 256 questions, with the largest domain (Literature, 43 questions) representing less than 17\% of the total. Full details of each stage appear in Appendix~\ref{app:filtering_algorithm}.

\subsection{Technical Evaluation Task}
\label{subsec:task_definition}

The course activity yields a technical evaluation task for expert-level deep research. Given a question $q$ requiring specialized domain knowledge and sophisticated retrieval strategies, a model $\mathcal{M}$ equipped with retrieval tools (web search, document visit) must produce an answer $\hat{a}$. The answer is evaluated against a reference answer $a$ using predefined grading criteria $\mathcal{G}$, yielding a score $s = \mathcal{G}(\hat{a}, a) \in [0, 100]$. All questions, reference answers, and grading criteria are in Chinese, and models must search and retrieve information from the open web in any language.

\textbf{Design Principles.} \MYBENCH~is designed around two principles that serve both evaluation and AI education:
\begin{itemize}[leftmargin=*,nosep]
    \item \textbf{Cross-Domain Coverage}: Questions span 14 normalized professional domains and 37 original field labels, including law, history, international relations, literature, foreign languages, social sciences, arts, archaeology, and political science. This diversity prevents models from relying on a single domain heuristic and lets students see that AI limitations are shaped by different knowledge structures.
    \item \textbf{Expert-Level Difficulty}: Each question represents a long-tail query requiring sophisticated retrieval strategies, deep domain knowledge, and careful interpretation of specialized sources. For students, this principle makes expertise visible: a hard question is bounded, verifiable, and resistant to shortcuts, rather than merely obscure.
\end{itemize}

These definitions and design principles connect the course activity to the model evaluation that follows. Section~\ref{sec:Evaluation} first summarizes the produced benchmark, then evaluates current deep research systems on it.

\section{Evaluation: Making AI Failures Observable}
\label{sec:Evaluation}

The evaluation is part of the educational design. It measures model performance, but it also tests whether student-designed tasks can surface failures that ordinary tool use may hide. A deep research system may search broadly, cite plausible sources, and produce a fluent answer while still asking the wrong query, trusting the wrong document, missing a disciplinary distinction, or extracting an imprecise answer. \MYBENCH~uses evaluation to make those failures inspectable for students and instructors. We first summarize the produced benchmark, then evaluate current systems and analyze where their work breaks down.

\subsection{Benchmark Overview}
\label{subsec:dataset_stats}

The course activity produces a benchmark with enough scale and diversity to support systematic evaluation. We report scale, domain coverage, difficulty, and answer characteristics because each dimension serves two purposes: it supports model evaluation and shows what kinds of disciplinary standards students turned into testable tasks for inspecting AI-mediated work.

\begin{figure}[H]
    \centering
    \includegraphics[width=0.45\linewidth]{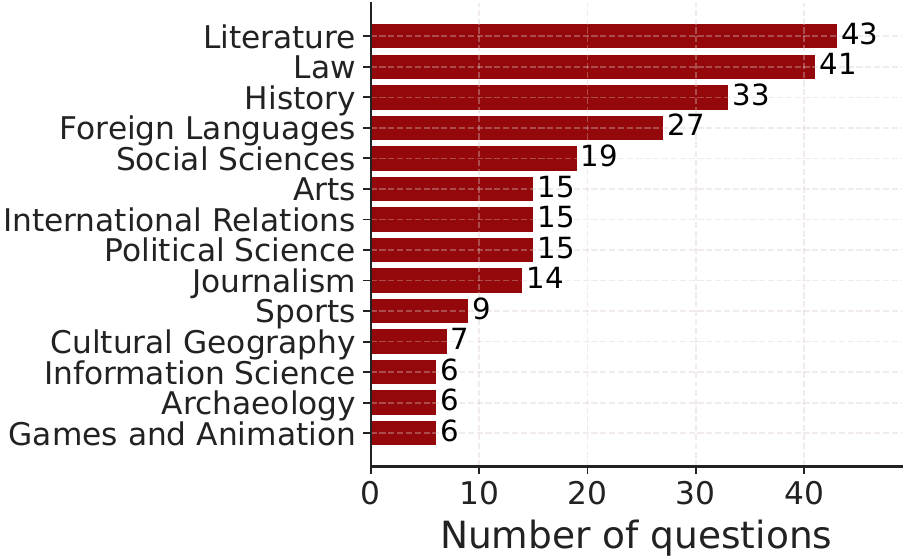}
    \includegraphics[width=0.54\linewidth]{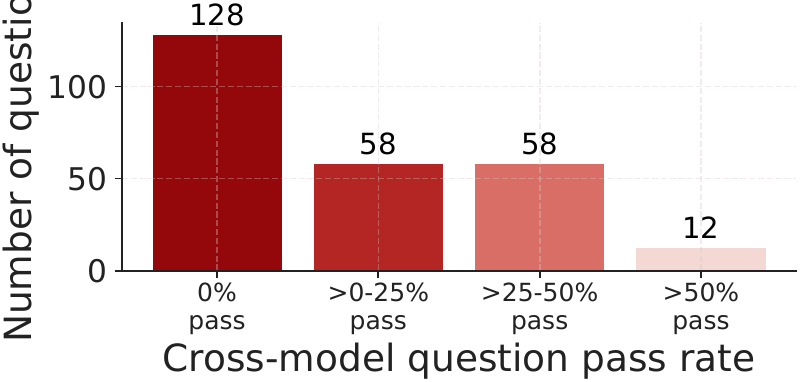}
    \caption{Left: Distribution across normalized domain groups. Right: Empirical cross-model question pass-rate distribution after evaluation.}
    \label{fig:distributions}
\end{figure}

\textbf{Scale and domain coverage.} \MYBENCH~comprises 256 questions spanning 14 normalized domains across the humanities and social sciences after normalization. Figure~\ref{fig:distributions} (left) visualizes the major domain groups, with the full normalized distribution reported in Appendix~\ref{app:dataset_stats}. Literature (43 questions), Law (41), History (33), Foreign Languages (27), and Social Sciences (19) form the largest categories, ensuring robust evaluation within these fields while maintaining broad cross-domain coverage.

\textbf{Difficulty distribution.} Questions are difficult enough to discriminate among frontier systems. Figure~\ref{fig:distributions} (right) shows the empirical cross-model pass-rate distribution after evaluation. Among 256 questions, \redtext{128} questions have 0\% cross-model pass rate among the four primary baselines, \redtext{58} questions have pass rate above 0\% and at most 25\%, \redtext{58} questions have pass rate above 25\% and at most 50\%, \redtext{and 12 questions exceed 50\% cross-model pass rate}. The average cross-model pass rate is \redtext{16.85\%} with median \redtext{3.85\%}, and the average cross-model score is \redtext{24.15}. The distribution supports the educational premise that students can construct tasks whose difficulty comes from disciplinary standards rather than hidden data or arbitrary trickiness.

\textbf{Question and answer characteristics.} Questions average 141.7 characters (range: 42--389). Answers average 29.3 characters (range: 1--455), indicating concise verifiable responses. Grading criteria average 73.6 characters and specify exact-match, partial-credit, or structured-answer requirements depending on the task. Detailed statistics including answer format distributions appear in Appendix~\ref{app:dataset_stats}.

\subsection{Evaluation Setup}
\label{subsec:eval_setup}

\textbf{Evaluated models.} We evaluate \redtext{thirteen} frontier search-enabled systems representing different model families and generations. \redtext{Four systems serve as primary baselines:} Kimi K2.5~\cite{kimiteam2026k25}, DeepSeek-V3.2~\cite{deepseekai2025v32}, Seed-2 Pro~\cite{bytedanceseed2026seed20}, and Seed-1.8 Pro~\cite{bytedanceseed2026seed18}. \redtext{The remaining nine systems are GPT-5.5~\cite{openai2026gpt55}, Claude Opus 4.7~\cite{anthropic2026opus47}, Gemini 3.1 Pro~\cite{google2026gemini31}, GLM 5.1~\cite{zhipu2026glm51}, DeepSeek-V4 Pro~\cite{deepseekai2026v4pro}, Kimi K2.6~\cite{kimiteam2026k26}, MiMo-V2.5 Pro~\cite{xiaomi2026mimo25}, Qwen 3.6 Plus~\cite{qwen2026qwen36}, and MiniMax M2.7~\cite{minimax2026m27}.} All models access identical \texttt{search} and \texttt{visit} tools with a nominal 50-call budget per question. Kimi K2.5 is evaluated with three runs per question (scores averaged within each question); all other models use one run. Scores normalize to 0--100, with 60 as the passing threshold. Full model configurations, prompt templates, and scoring details appear in Appendix~\ref{app:eval_details}.

\subsection{Do Student-Designed Tasks Make AI-Mediated Work Inspectable?}
\label{subsec:main_results}

The overall scores show that student-designed tasks make expert-level reliability visible as a problem, even for strong deep research systems. Table~\ref{tab:main_results} presents overall results. \redtext{GPT-5.5 leads with the highest mean score (67.12) and pass rate (57.58\%), followed by Claude Opus 4.7 (57.79, 52.94\%) and Gemini 3.1 Pro (43.07, 31.82\%). Among the four primary baselines, Kimi K2.5~\cite{kimiteam2026k25} achieves the highest mean score (30.48) and pass rate (25.26\%), followed by Seed-2 Pro~\cite{bytedanceseed2026seed20} (22.89, 15.23\%), Seed-1.8 Pro~\cite{bytedanceseed2026seed18} (22.07, 12.50\%), and DeepSeek-V3.2~\cite{deepseekai2025v32} (14.58, 7.81\%).} The median score is 0 for all primary baseline models except Kimi K2.5, whose median question-level score is 16.67. Current systems can solve some questions, but they cannot be treated as reliable workers for expert-level cross-domain retrieval without further inspection.

For the paper's educational argument, difficulty matters because it identifies which parts of the work require accountable judgment. Reliable answers require discipline-specific ways of asking, locating, interpreting, and verifying. The questions are publicly answerable, but they cannot be solved by fluent synthesis alone. In the course, the score table becomes a starting point for asking where the work broke down: the task definition, the query, the source path, the evidence, or the final acceptance of an answer.

\begin{table}[t]
\centering
\caption{Overall model performance on \MYBENCH. Pass rate uses the 60-point threshold. \redtext{Kimi K2.5 scores are averaged over three runs per question.}}
\label{tab:main_results}
\redtext{
\small
\begin{tabular}{lccc}
\toprule
Model & Mean Score (0--100) & Pass Rate (\%) & Avg. Tool Calls \\
\midrule
GPT-5.5 & 67.12 & 57.58 & 41.12 \\
Claude Opus 4.7 & 57.79 & 52.94 & 41.94 \\
Gemini 3.1 Pro & 43.07 & 31.82 & 18.34 \\
GLM 5.1 & 42.30 & 35.14 & 61.70 \\
DeepSeek-V4 Pro & 38.24 & 32.43 & 23.92 \\
Kimi K2.6 & 37.74 & 32.26 & 30.81 \\
Qwen 3.6 Plus & 36.05 & 30.95 & 45.83 \\
MiMo-V2.5 Pro & 32.05 & 22.73 & 40.68 \\
Kimi K2.5 & 30.48 & 25.26 & 33.41 \\
MiniMax M2.7 & 29.29 & 16.67 & 32.38 \\
Seed-2 Pro & 22.89 & 15.23 & 14.33 \\
Seed-1.8 Pro & 22.07 & 12.50 & 39.71 \\
DeepSeek-V3.2 & 14.58 & 7.81 & 24.76 \\
\bottomrule
\end{tabular}
}
\end{table}

\begin{figure}[t]
    \centering
    \includegraphics[width=0.475\linewidth]{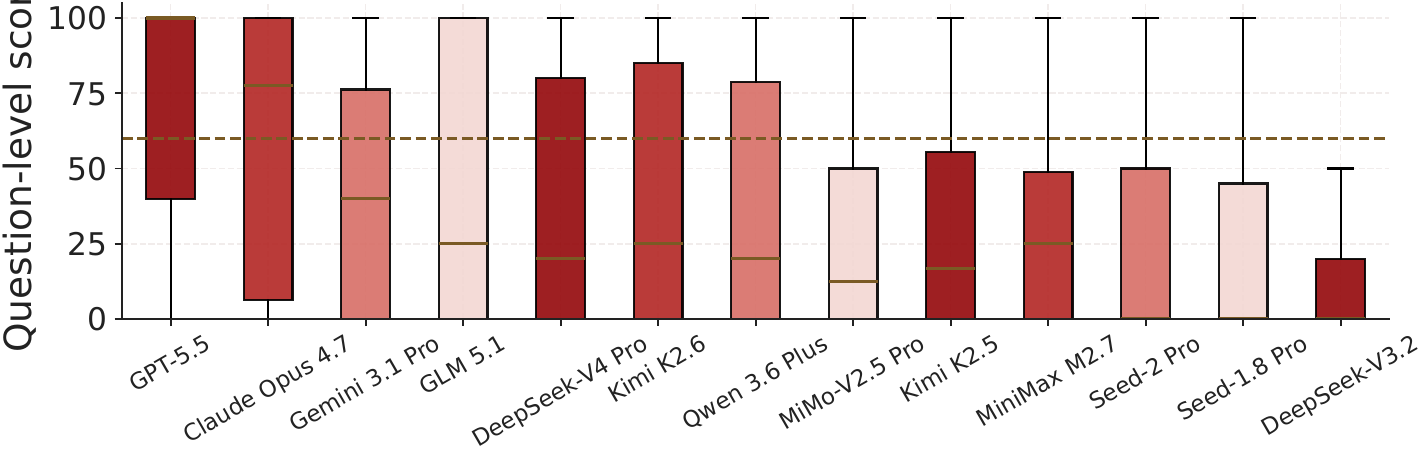}
    \includegraphics[width=0.475\linewidth]{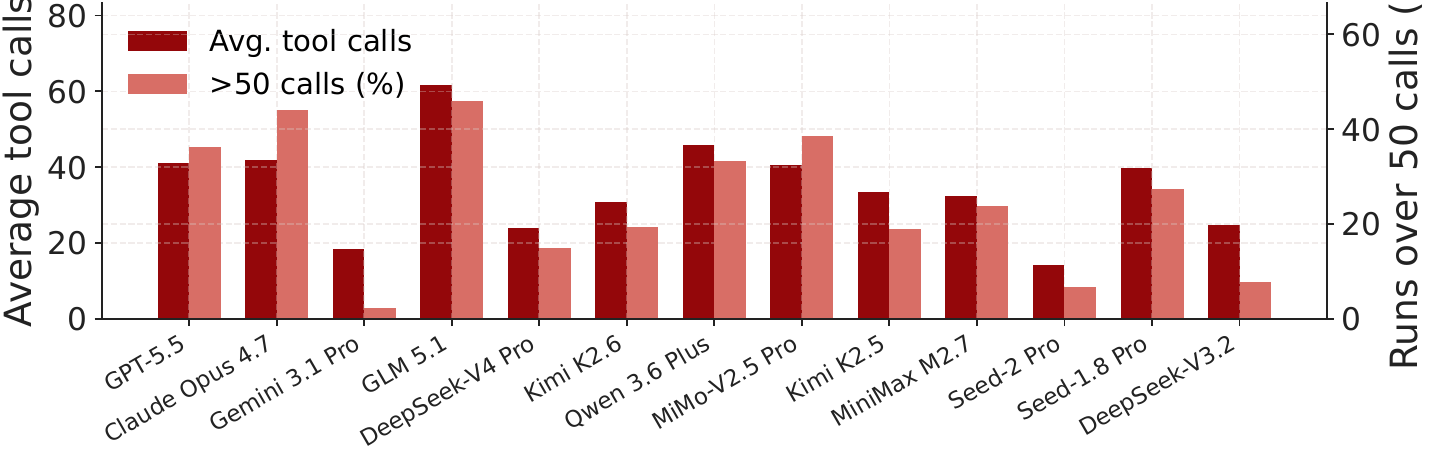}
    \caption{Left: score distributions across models. Right: average tool calls and fraction of runs exceeding 50 calls. \redtext{Among the thirteen evaluated models, higher tool use does not consistently correspond to higher accuracy: GLM 5.1 averages 61.70 calls but ranks fourth, while Gemini 3.1 Pro reaches third place with only 18.34 calls.}}
    \label{fig:score_tool_analysis}
\end{figure}

Domain-wise performance shows that AI reliability depends on disciplinary context. Performance remains low across domains, so the benchmark tests broad capabilities rather than one narrow knowledge area. \redtext{Averaged across thirteen models, Social Sciences (36.83 mean, 19 questions) and Foreign Languages (32.70, 27 questions) are the highest-scoring sizeable domains, while Sports (12.28, 9 questions) and International Relations (16.37, 15 questions) are the most difficult.} Rankings for small domains should be interpreted cautiously. The cross-domain pattern shows that reliability is not a single global property of a model. It changes with source ecosystems, disciplinary terminology, and answer standards. Students therefore need more than a general habit of checking AI. They need standards from the field in which the answer will be used.

Question difficulty also supplies material for practicing inspection. \redtext{The majority of questions remain unsolved by all evaluated models, and only 12 exceed 50\% cross-model pass rate}. The mean question-level pass rate is \redtext{16.85\%} (median \redtext{3.85\%}), rising to \redtext{23.84\%} after excluding all-zero-score questions. These numbers show that \MYBENCH~discriminates among model capabilities while leaving substantial headroom. In a course setting, they provide concrete cases for analyzing which task was defined, which source was trusted, which evidence was missing, and what a responsible user would have had to decide before using the output.

\subsection{Where Does the Work Break Down?}
\label{subsec:analysis}

\begin{figure}[t]
    \centering
    \begin{subfigure}[t]{0.6\linewidth}
        \centering
        \includegraphics[width=\linewidth]{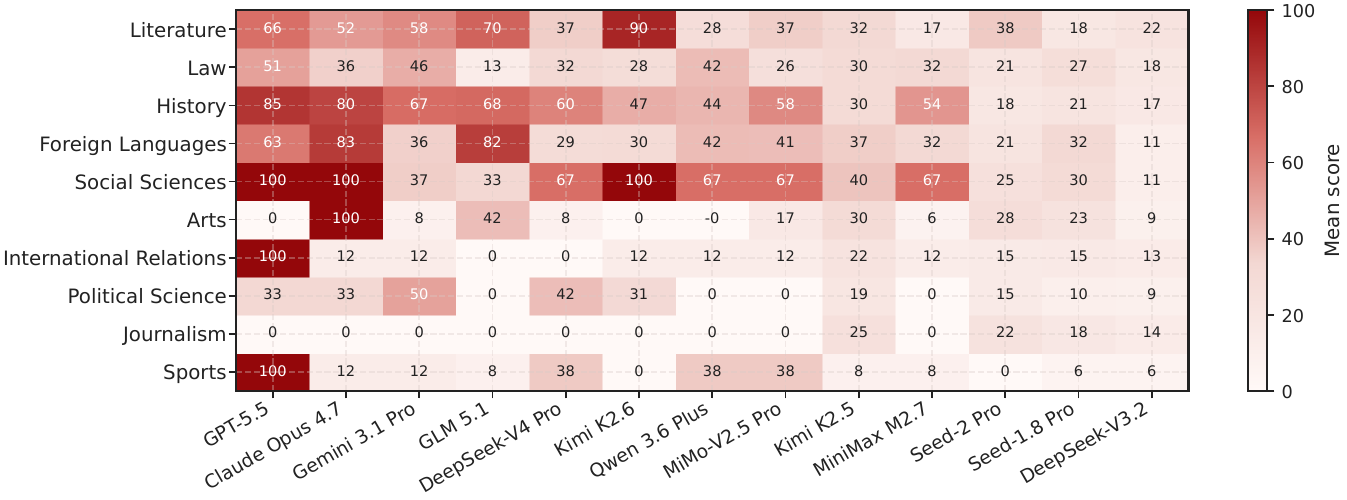}
        \caption{Domain-wise mean scores.}
        \label{fig:domain_performance}
    \end{subfigure}\hfill
    \begin{subfigure}[t]{0.38\linewidth}
        \centering
        \includegraphics[width=\linewidth]{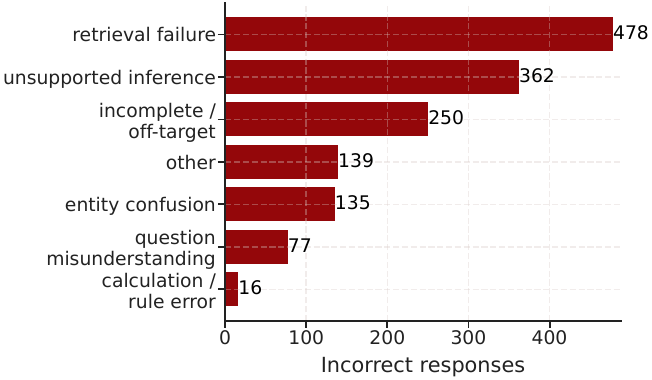}
        \caption{Failure pattern counts.}
        \label{fig:failure_patterns}
    \end{subfigure}
    \caption{Domain and error analysis. Left: mean scores for the largest normalized domains \redtext{across the thirteen evaluated models}. Right: failure category distribution across 1,457 non-passing responses \redtext{from the four primary baseline models}.}
    \label{fig:domain_error_analysis}
\end{figure}

Failure analysis explains what the scores mean. We conduct a systematic error analysis on all non-passing responses across \redtext{the four primary baseline models (Kimi K2.5, DeepSeek-V3.2, Seed-2 Pro, Seed-1.8 Pro). The error-pattern statistics below cover these four baselines; failure analysis for the remaining nine systems is left to future work.} Each response that scores below 60 is classified into one of seven failure categories by an annotator who reads the model's search trace and final answer. The analysis covers 1,457 responses in total (723 from Kimi K2.5 across three runs, and approximately 248 each from the other three models). Figure~\ref{fig:failure_patterns} shows the distribution.

Naming failure types turns the vague impression that ``AI is wrong'' into an inspection of the work process. Retrieval failure is the most frequent category (478 responses, 32.8\%): models cannot locate relevant information despite its public availability, especially when information is embedded in specialized archives, government documents, library metadata, or non-English sources. Unsupported inference is the second major pattern (362, 24.8\%): when search evidence is incomplete, models infer a plausible answer from partial matches rather than continuing verification. Incomplete or off-target answering accounts for 17.2\% (250 responses), where models retrieve useful context but fail to output the exact entity, number, or legal term required by the grading criteria. Entity confusion (135, 9.3\%) and question misunderstanding (77, 5.3\%) are common when multiple similar documents, artworks, statutes, or historical figures satisfy nearby constraints. Calculation or rule errors are rare (16, 1.1\%). The remaining 139 responses (9.5\%) involve timeout or empty output, where models exhausted their tool budget or failed to produce a final answer.

These failure patterns map onto the work students have to learn to supervise. \textit{Query formulation} failures (retrieval failure and question misunderstanding, 38.1\% combined) indicate that models cannot construct domain-appropriate searches when general-purpose query strategies fail. \textit{Source navigation} failures (entity confusion, 9.3\%) show difficulty distinguishing closely related documents or entities within structured professional archives. \textit{Answer extraction} failures (unsupported inference, incomplete answering, and calculation errors, 43.1\% combined) reveal that even when relevant sources are found, models struggle to produce the precise output that expert-level grading criteria demand. These are also the points where students must exercise judgment: define the right query, know which source identity matters, and decide whether an answer satisfies the field's standard.

\textbf{Representative examples.} We present three illustrative cases:

\textit{Case 1 (Answer extraction -- Law, terminology precision):} A question asks for a civil-law institution that transforms a three-party legal relationship into a relationship involving four or more parties, then asks for the crime in the Criminal Code with the same article number as the corresponding Civil Code provision. The correct answer is the crime of low-price share discounting and sale of state-owned assets for personal gain. Models often identify the civil-law institution and the relevant article number, but fail because they output a nearby or updated crime name rather than the exact statutory formulation required by the grading criteria. Legal search in this case requires precise tracking of specialized terminology, not just conceptual understanding.

\textit{Case 2 (Source navigation -- International Relations, archival failure):} A question requires identifying the FRUS document number for ABC reporter John Scali's October 26, 1962 meeting with Soviet KGB officer Aleksandr Fomin at the Occidental Restaurant. Models often navigate to the correct FRUS volume and locate nearby documents involving Scali and Fomin, but select a related memorandum rather than the exact document number. The error is a structured archival navigation failure: relevant evidence is available, but precise metadata and document-level distinctions remain difficult.

\textit{Case 3 (Query formulation -- Arts, last-hop trap):} A question requires a three-hop search: identify a monograph on Spanish visual culture from a quoted passage, find its dedicatee, then determine which film is analyzed in Chapter 4 of the dedicatee's 2002 Spanish-language book. All models complete the first two hops (identifying the book and dedicatee) but fail at the final hop, where the target is a specialized academic monograph with minimal web indexing. Multi-hop search difficulty is not linear: it increases sharply when the chain enters low-indexing specialized sources, and models lack the ability to judge when a search path has moved beyond standard web coverage.

These examples show that failure rarely comes from information unavailability alone. It comes from the interaction of specialized query formulation, precise source navigation, and strict answer extraction. In the course setting, such failures are useful because they make AI-mediated work inspectable. Students can see that a model may reach the right source family but miss the exact document, identify the right legal concept but output the wrong statutory version, or complete early search hops while failing at the least-indexed disciplinary source. These cases show professional judgment at work in the evaluation process. Additional detailed case studies with full model outputs are provided in Appendix~\ref{app:case_studies}.

\textbf{Computational costs.} More search does not reliably improve performance. \redtext{GLM 5.1 averages 61.70 tool calls per question but ranks fourth, while Gemini 3.1 Pro reaches third place with only 18.34 calls. The two top performers, GPT-5.5 and Claude Opus 4.7, use moderate budgets (41.12 and 41.94 calls).} Effective expert search requires query quality and source judgment, not just search depth (see Appendix~\ref{app:eval_details} for full tool-use analysis). This result supports the paper's educational claim in a concrete way: using a stronger or more active tool is not enough unless the tool's activity is guided and evaluated by meaningful human standards.

\section{Discussion: What Benchmark Construction Teaches About AI Education}
\label{sec:Discussion}

\MYBENCH~uses benchmark construction to connect tool exposure with responsibility for AI-mediated knowledge work. Students first encounter deep research as a real AI productivity tool. They then turn the tool into something they can inspect by writing questions, grading criteria, source justifications, and anti-shortcut checks. Deep research remains the concrete case, but the educational issue is broader: students need to learn how to define, guide, and judge work in which AI systems search, read, write, and use tools. The activity does not ask students to stop making and become detached judges. It asks them to make the task, the standard, and the inspection procedure that determine whether an AI-produced answer can count as knowledge.

\subsection{Tool Exposure Still Matters}

Direct exposure matters. Students cannot learn AI only through abstract discussion; they need to see what current systems can actually do. Deep research systems are useful for this purpose because they feel like real productivity tools. They search, synthesize, and produce answers with sources. This first layer puts students inside the kind of AI-mediated knowledge environment they are likely to inhabit.

The course then changes the student's relation to the tool. A productive tool is not automatically a reliable authority. A fluent answer with sources may appear to have transformed information into knowledge, but the evaluation results show why that appearance can be unstable. Knowledge still depends on the value of the question, the authority of the source, the meaning of terms within a discipline, and the person who decides whether the answer is ready to use.

\subsection{Task Design as Accountable Work}

Much AI education teaches students how to ask systems better questions. Benchmark construction asks them to design questions that can test a system. That difference matters. A prompt can be improved by changing wording. A benchmark question has to define a problem space: the intended reasoning path, the evidence needed to verify the answer, the shortcuts that must be blocked, and the grading standard that separates a correct answer from a plausible one.

This is where the activity adds a second layer to ordinary tool use. Students still use AI, but they also construct the conditions under which its work can be judged. They cannot outsource the structure of the task to the model, because the task is what they are building. They have to decide what is worth asking and how an answer should be checked. In AI-mediated knowledge work, that kind of problem formulation is not a preliminary step before real work begins. It is part of the work.

\subsection{Disciplinary Knowledge as Evaluative Authority}

\MYBENCH~also gives disciplinary knowledge a concrete role. In the presence of strong AI systems, students may wonder whether professional knowledge matters less because models can retrieve and summarize information on demand. The construction process points the other way. Professional knowledge tells students what to ask, which sources matter, what distinctions cannot be collapsed, and why a fluent answer may still be wrong.

Attention to evidence, sources, terminology, and precision should therefore not be treated as a set of mechanical behaviors. They are ways in which a field makes knowledge accountable. Evidence links a claim to something that can be checked. Source identity connects an answer to a knowledge community. Terminology preserves conceptual boundaries. Precision matters because professional work often fails at the level of the exact article number, document identifier, edition, translation, or statutory name. Benchmark construction makes these standards visible because students must write them down and defend them before models are evaluated.

\subsection{Student Reflections and the Shift in Role}

Reflections from five student contributors help ground the educational claim. Their role here is not to settle the long-term impact of the course, but to show what kinds of change become visible in this setting. Students described deep research systems less as neutral answer providers and more as tools whose outputs had to be situated and tested. Several reflections returned to the same practical discovery: designing a good question could be harder than obtaining an answer, because the question had to make the evidence path, answer boundary, and grading standard explicit.

That discovery changes what AI use means for the student. If the model can search, read, write, code, and call tools, the student's role does not disappear. It expands across the whole chain of work. Students still make things: questions, evidence paths, standards, rubrics, and judgments about use. The change is that making and judging become harder to separate. Benchmark construction can therefore cultivate epistemic agency in a precise sense. Students learn to remain responsible knowledge actors even when machines participate in producing answers.

\subsection{Reusable Practice and Long-Term Educational Inquiry}

The course activity behind \MYBENCH~can be reused beyond this benchmark. A course with students from different majors can ask them to contribute candidate questions from their own fields, review each other's designs, validate answers, and evaluate AI systems on the resulting tasks. The goal need not be a public benchmark in every setting. Even when the primary goal is educational, the same structure can help students examine the relationship among AI tools, disciplinary knowledge, evidence standards, and human responsibility.

The larger question is how education should respond as AI becomes ordinary in learning and professional work. The value of practices such as benchmark construction is unlikely to be captured by short-term performance measures alone. What matters is whether students develop the habits needed to use AI without surrendering the work of judgment. It also matters whether they see professional knowledge as a source of responsibility rather than a body of content to be automated away. Answering that question requires longer-term observation across courses, cohorts, disciplines, and forms of evidence, including reflective interviews, classroom observations, student artifacts, and follow-up studies of how students use AI in later work.

\MYBENCH~is one setting in which that inquiry can be pursued. It makes AI-mediated knowledge work observable in the classroom. Students confront the gap between fluent output and trustworthy knowledge, between using a tool and accepting responsibility for an answer, and between professional knowledge as content and professional knowledge as a standard for evaluating AI. How these experiences accumulate into durable intellectual habits remains open. The contribution of \MYBENCH~is to make that question concrete enough to study and refine.

\section{Conclusion}
\label{sec:Conclusion}

This paper presents \MYBENCH~as a course-based practice for teaching accountable AI-mediated knowledge work through benchmark construction. The paper contributes a set of difficult questions for deep research systems and a course structure in which students learn AI by building the tests themselves. Students first encounter deep research as an AI-era productivity tool. They then construct expert-level tasks, review those tasks under peer scrutiny, evaluate models, and use failures as evidence for discussing what trustworthy knowledge requires.

The benchmark artifact remains important. \MYBENCH~contains 256 expert-level questions spanning 14 professional domains, constructed through course-based collaboration with 85 student question authors. Evaluation of thirteen frontier deep search systems shows clear limitations: the best-performing system achieves only 57.58\% pass rate, and many questions remain unsolved by every evaluated system. Failure analysis identifies three interacting bottlenecks: query formulation failures (38.1\%), source navigation failures (9.3\%), and answer extraction failures (43.1\%). These results show that student-designed tasks can reveal where AI-mediated work breaks down. For the paper's educational argument, those failures become material for a larger lesson: reliable knowledge depends on more than fluent answer generation.

\MYBENCH~suggests that AI education should develop more than efficient tool users. As AI systems become more capable, students also need practice in the parts of work that cannot be treated as automatic: framing meaningful questions, setting evidence standards, inspecting process, and deciding when an answer is reliable enough to use. Benchmark construction makes this lesson concrete. It helps students encounter AI as a knowledge-work tool, understand it as a system with reliability limits, use model failures as material for reflection, and define their own role in AI-mediated work. The benchmark is therefore more than an evaluation artifact. It is a way to ask what kind of judgment education should cultivate when AI participates in knowledge production.

\textbf{Responsible use.} The benchmark should be used to examine AI-mediated knowledge work, not as a fixed ranking of systems or a replacement for course design. Model scores will change as tools, search APIs, and web coverage change. The more durable value is the activity of making failures inspectable: asking which question was defined, which source was trusted, which evidence was used, and who remains responsible for the final judgment. In educational settings, the activity should protect student privacy and avoid using benchmark performance as a proxy for student ability. The goal is to help students develop judgment, not to turn their disciplinary work into a narrow contest of model scores.

\textbf{Limitations and future inquiry.} The limitations of \MYBENCH~follow from its dual role as a research artifact and a course-based activity. First, the benchmark currently covers humanities and social science domains and does not include natural sciences, engineering, or medical fields, where expert search may involve formulas, experimental data, instruments, or safety-critical evidence. Second, the construction process depends on students with advanced disciplinary training and on instructors who can support multi-round review, so the activity may need adaptation in other institutions, course formats, or student populations. Third, the educational impact of AI is itself a long-term and still-unfolding question. Reflections from five student contributors help identify what is worth observing, but the development of judgment, responsibility, and epistemic agency should be studied across courses, cohorts, disciplines, and later uses of AI. Fourth, the model evaluation reflects a specific snapshot of deep search systems, tool interfaces, and open-web availability. Most systems are evaluated with one run per question, so the reported scores should be read as descriptive evidence of current failure patterns rather than as statistically stable estimates of permanent model rankings. These limits define the scope of the paper's claim. \MYBENCH~proposes and studies a course-based benchmark construction activity through which students can encounter AI tools, construct verifiable problems, inspect model failures, and practice the forms of judgment that AI-mediated knowledge work requires.

\begin{ack}

We thank the Peking University students who participated in the course-based construction of \MYBENCH. Their work on question design, adversarial review, grading criteria, and answer validation made the benchmark possible and shaped the educational practice studied in this paper. To reflect both the final benchmark and the collaborative construction process, we acknowledge the contributors below.

\textbf{Benchmark construction contributors (101).} Dama Ba, Yang Bai, Jiyue Cao, Lijun Chen, Sifan Chen, Siman Chen, Siqi Chen, Yaoyao Chen, Yujie Chen, Yuqi Chen, Qi Cui, Tianyu Dai, Wenyi Dai, Yuanyuan Deng, Haotian Ding, Sihan Du, Chengyu Fan, Dexin Fei, Zichang Gao, Jiayun Gu, Xinqing Hao, Ruoyi Hu, Weiyu Hu, Yajie Hu, Dingyi Huang, Rensong Huang, Shucheng Huang, Yining Huang, Yutong Jiang, Xiaotong Jiao, Congyu Jie, Xiaozhen Jin, Yedi Jin, Menghan Li, Xinyuan Li, Ruikang Lin, Yihang Lin, Kewei Liu, Luoyirui Liu, Shijing Liu, Sijia Liu, Siyu Liu, Tianqi Liu, Xuanyu Liu, Yawen Liu, Linyong Long, Yuqi Lu, Xianchen Meng, Yufan Niu, Shuyan Pu, Yifan Shen, Yihe Shi, Wei Song, Yuting Sun, Yitong Tan, Yujiao Tao, Dijing Wang, Jiayi Wang, Yanzhi Wang, Yuhao Wang, Jiayi Wei, Peiru Wei, Qiulin Wei, Yuxin Wei, Xinyi Weng, Gulijiangbali Wu, Yawen Xiao, Bingxin Xie, Shurui Xie, Zhi Xing, Li Ya, Yichen Yan, Hanqing Yang, Qingyue Yang, Guanzhi Yu, Tongze Yuan, Siyu Zhang, Yihan Zhang, Yumeng Zhang, Yuxuan Zhang, Jingwen Zhao, Jiachen Zheng, Shengnan Zheng, Xiaoyu Zheng, Youzhen Zheng, Chaoran Guo, Jiayan Guo, Xinyue He, Binzhou Li, Rongcai Li, Shuoren Liu, Zhengda Lu, Xin Shu, Karen Uchibori, Zhengyi Wang, Zhichu Wu, Mu Xiong, Wenxiao Xiong, Zhizheng Yang, Zhizhi Zhang, Bozhong Zhao.

We also thank the course instructors and teaching staff who facilitated the collaboration and supported the multi-round review process. We acknowledge the computational and construction support used to run model evaluations and organize the resulting analyses.

\end{ack}

\bibliographystyle{plainnat}
\bibliography{reference}

\appendix

\section{Course Protocol for Benchmark Construction}
\label{app:design_protocol}

This appendix gives the question design protocol used in the course-based benchmark construction activity. The protocol keeps questions expert-level, verifiable, and resistant to common shortcuts. It also shows how the course asks students to define problems, defend evidence, specify what would make an answer trustworthy, and inspect the AI-mediated work that produces it.

\subsection{Core Requirements}

Each question must satisfy the following requirements:

\textbf{Domain Expertise Requirement.} Questions must require specialized knowledge from the designer's professional field that would not be accessible to educated non-experts. The question should resemble a research scenario or professional task that domain practitioners would recognize as challenging.

\textbf{Long-tail Information Targeting.} Questions should target information that is publicly available but not widely known, embedded in sources such as academic publications, professional archives, or domain-specific databases. Effective searching requires domain knowledge to formulate appropriate queries and interpret specialized terminology.

\textbf{Answer Uniqueness and Verifiability.} Provide a unique, objectively correct answer (or well-defined set of correct answers) with clear verification method through authoritative sources. Acceptable formats include single entities, entity lists, structured objects, ordered sequences, tables, or numerical results. All possible correct answers must be enumerable and specified.

\textbf{Detailed Grading Criteria.} Specify point allocation for each component of the answer. Include rules for partial credit, alternative phrasings, and additional or missing information. Criteria must support objective evaluation, appropriate partial credit, and edge-case handling.

\textbf{Construction Documentation.} Document the evolution from a simple question to an expert-level query, explaining what each iteration adds and why the final version requires domain expertise. This record helps students see that problem formulation is part of accountable AI-mediated knowledge work, not a preliminary step before it.

\subsection{Anti-Shortcut Checklist}

Before submission, designers verify that questions avoid three common shortcut traps:

\textbf{Trap: Pre-existing Solutions.} Search combinations of question terms in multiple search engines to verify that no webpage directly provides the answer with the same reasoning path. If such a page exists, reformulate the question to require different information integration or add constraints that require independent synthesis.

\textbf{Trap: Trivial Identifiers.} Ensure that no single constraint acts as a unique identifier (ISBN, DOI, atomic number, or a specific date paired with a name) that determines the answer without reasoning. Replace unique identifiers with combinations of non-unique descriptive properties.

\textbf{Trap: Bypassable Multi-step Design.} For questions that appear to require multiple steps, verify that those steps are necessary and cannot be bypassed through direct search. Test whether combinations of start and end constraints yield the answer without intermediate reasoning. If they do, strengthen the dependencies between steps or add constraints that require information synthesis.

\section{Peer Review and Filtering Details}
\label{app:filtering_algorithm}

Algorithm~\ref{alg:iterative_filtering} in Section~\ref{subsubsec:iterative_filtering} gives the filtering workflow. We add details for each stage here.

\textbf{Stage 1 details.} Preliminary screening removes questions with the following issues: (a) questions requiring access to private or paywalled content that cannot be verified through open-web search; (b) ambiguous wording where multiple reasonable interpretations lead to different answers; (c) questions directly answerable by the baseline model with high confidence, indicating insufficient difficulty; and (d) questions where no evaluated model produces a meaningful search attempt, suggesting potentially ill-defined requirements.

\textbf{Stage 2 details.} In the iterative validation rounds, each reviewer works independently without seeing other reviewers' assessments. For Round 1 (answer correctness), reviewers must independently locate authoritative sources confirming the reference answer and verify that no alternative correct answers exist. For Round 2 (grading criteria audit), reviewers simulate scoring sample responses to verify that criteria produce consistent scores. For Round 3 (anti-shortcut verification), reviewers test at least three different search query formulations to confirm that no straightforward search path yields the answer directly. These validation steps also serve the educational goal: students learn that accountable AI-mediated work depends on standards that can be inspected and defended by others.

\textbf{Disagreement resolution.} When reviewers disagree, the question is returned to the original author with specific feedback. If disagreement persists after revision, the question is removed from the candidate pool. This conservative rule prioritizes benchmark quality over quantity. It also models an important norm for AI education: knowledge claims should survive scrutiny rather than rely on private confidence or fluent presentation.

\section{AI Failure Cases for Classroom Analysis}
\label{app:case_studies}

This section presents detailed cases from \MYBENCH~evaluation, with question specifications, model responses, and analysis. They document where AI-mediated knowledge work breaks down and provide classroom material for discussing why such work requires domain standards and human judgment. These cases correspond to the representative examples summarized in Section~\ref{subsec:analysis}.

\subsection{Case 1: Legal Domain -- Answer Extraction Failure}

\textbf{Domain}: Law \quad \textbf{Challenge dimension}: Answer extraction (terminology precision)

\textbf{Question} (translated from Chinese): \textit{There is a civil-law institution that transforms a legal relationship originally occurring among three parties into one involving four or more parties. This type of legal act often involves contracts but does not appear in the Contract Section of the Civil Code. State the crime in the Criminal Code whose article number matches that of the Civil Code provision governing this institution.}

\textbf{Reference answer}: The crime of personal-gain-motivated low-price share discounting and sale of state-owned assets (translated from Chinese), Article 169.

\textbf{Grading criteria}: 100 points for an exact character-by-character match of the crime name; 0 otherwise.

\textbf{Reasoning path}: The civil-law institution is \textit{sub-agency} or entrusted sub-agency (translated from Chinese), governed by Article 169 of the Civil Code (General Provisions, Chapter 7: Agency). Sub-agency extends the original three-party relationship (principal, agent, counterparty) to four or more parties by introducing a sub-agent. The corresponding crime in Article 169 of the Criminal Code is the target answer.

\textbf{Model responses}:

\begin{itemize}[nosep]
\item \textbf{Seed-2 Pro} (0/100): Incorrectly identified the civil-law institution as ``counter-guarantee'' (Article 387), leading to the wrong Criminal Code article and the answer ``crime of unit bribery.''
\item \textbf{Seed-1.8 Pro} (0/100): Correctly identified sub-agency and Article 169, but answered ``crime of personal-gain-motivated low-price share discounting and sale of company or enterprise assets'' (translated from Chinese), which is the post-2024-amendment name, not the original statutory formulation.
\item \textbf{Kimi K2.5} (run 1: 0/100, run 2: 100/100, run 3: 100/100): In the failing run, Kimi produced the post-amendment name (same error as Seed-1.8). In two successful runs, it provided the exact pre-amendment statutory formulation.
\item \textbf{DeepSeek-V3.2} (0/100): Produced no answer (empty response).
\end{itemize}

\textbf{Analysis}: This case contains two answer extraction failures. Seed-2 Pro failed at the initial reasoning step by misidentifying the civil-law institution. Seed-1.8 Pro and one Kimi run completed the reasoning path but failed at final answer extraction because of a terminology-evolution trap: the 2024 Criminal Code amendment expanded the crime's scope from ``state-owned assets'' to ``company or enterprise assets,'' changing the official name. Models that retrieved post-amendment sources produced the updated name, which does not match the grading criteria requiring the original statutory formulation. Expert-level legal tasks require precise tracking of legislative versioning, not just conceptual understanding.

\subsection{Case 2: International Relations -- Source Navigation Failure}

\textbf{Domain}: International Relations \quad \textbf{Challenge dimension}: Source navigation (archival metadata precision)

\textbf{Question} (translated from Chinese): \textit{Consult the Foreign Relations of the United States (FRUS) volume on the Cuban Missile Crisis (Volume XI, 1961--1963). Locate the document number of the memorandum in which ABC reporter John Scali reported to the State Department about his October 26, 1962 meeting with Soviet KGB officer Aleksandr Fomin at the Occidental Restaurant in Washington, DC.}

\textbf{Reference answer}: Document 64.

\textbf{Grading criteria}: 100 points for ``Document 64'' or ``64''; 0 for any other document number.

\textbf{Model responses}:

\begin{itemize}[nosep]
\item \textbf{Seed-2 Pro} (0/100): Answered ``Document 85,'' described as an editorial note.
\item \textbf{Seed-1.8 Pro} (0/100): Answered ``Document 80,'' identified a memorandum titled ``Memorandum From ABC Correspondent John Scali to the Director of the Bureau of Intelligence and Research.''
\item \textbf{Kimi K2.5} (run 1: 0/100, run 2: 0/100, run 3: 0/100): Answered ``Document 195,'' ``Document 80,'' and ``Document 195'' across three runs. None matched.
\item \textbf{DeepSeek-V3.2} (0/100): Answered ``Document 80.''
\end{itemize}

\textbf{Analysis}: All models failed this question across all runs, making it one of the universally unsolved questions. Three of four models, and two of three Kimi runs, converged on Document 80. That is a real FRUS document involving Scali, but it records a \textit{different} communication in the sequence. FRUS Volume XI contains dozens of documents from overlapping dates with the same participants (Scali, Fomin, Hilsman). Document 64 records the initial restaurant meeting, while Documents 80 and 195 record later follow-up communications. Models consistently located the correct volume, date, and participants, but could not distinguish between closely related documents in a structured archive. Professional archival navigation requires content-level analysis of individual documents rather than metadata-level matching.

\subsection{Case 3: Cross-Domain Scholarly Search -- Query Formulation Failure}

\textbf{Domain}: Arts \quad \textbf{Challenge dimension}: Query formulation (multi-hop search)

\textbf{Question} (translated from Chinese): \textit{A scholarly monograph on contemporary Spanish visual culture analyzes Bleda y Rosa's photographic work ``Hacia Valeria.'' The author notes that ``Hacia'' (Towards) suggests we are placed ``in medias nusquam,'' never truly reaching history through images. The author dedicated this book to a female scholar who deeply influenced her. Find this dedicatee, then identify: in that scholar's 2002 Spanish-language monograph on ``wounded culture'' (cultura herida), which director's which film is primarily analyzed in Chapter 4?}

\textbf{Reference answer}: Director Iv\'{a}n Zulueta's film \textit{Arrebato}. The reasoning chain: Book A = Patricia M. Keller, \textit{Ghostly Landscapes} $\rightarrow$ dedicatee = Cristina Moreiras-Menor $\rightarrow$ Book B = \textit{Cultura herida} (2002) $\rightarrow$ Chapter 4 analyzes \textit{Arrebato}.

\textbf{Grading criteria}: Path verification (40 points): correctly identify Keller's book and Moreiras-Menor as dedicatee. Final answer (60 points): Iv\'{a}n Zulueta (30 points) + \textit{Arrebato} (30 points).

\textbf{Model responses}:

\begin{itemize}[nosep]
\item \textbf{Seed-2 Pro} (40/100): Correctly identified Keller's book and Moreiras-Menor, but gave the wrong film for Chapter 4.
\item \textbf{Seed-1.8 Pro} (20/100): Identified Moreiras-Menor but attributed the wrong chapter content.
\item \textbf{Kimi K2.5} (run 1: 40, run 2: 40, run 3: 40): All three runs correctly completed the first two hops (Keller $\rightarrow$ Moreiras-Menor) but failed to identify the correct Chapter 4 content in \textit{Cultura herida}.
\item \textbf{DeepSeek-V3.2} (40/100): Same pattern: correct path verification, incorrect final answer.
\end{itemize}

\textbf{Analysis}: This question requires a three-hop search chain across two books and two scholars. All models completed the first two hops: identifying Keller's monograph from the Bleda y Rosa analysis, then finding Moreiras-Menor as the dedicatee. The failure occurs at the final hop, identifying the content of Chapter 4 in Moreiras-Menor's 2002 Spanish-language monograph. The book is a specialized academic work with limited online presence, and its table of contents is not readily available through standard web search. The consistent partial success (40/100 across most models) followed by final-hop failure shows that multi-hop expert search becomes harder as the chain enters less-indexed specialized sources.

\section{Example Student-Constructed Questions}
\label{app:example_questions}

This section gives representative questions from different domains, including grading criteria and construction rationale. These examples show what it means for students to turn disciplinary knowledge into a verifiable AI evaluation task.

\subsection{Example from Foreign Languages and Cultural Studies}

\textbf{Question}: \textit{Please help me find a fairy tale written between the year when World War II reached its largest scale and the year when the second front of World War II was opened. The author's country participated in both World War I and World War II, and the author had a profession other than writer. The fairy tale was adapted into both a film and a musical. Find the most common public Italian version of this fairy tale, identify the most frequent word in Chapter 6, and report its frequency.} (translated from Chinese)

\textbf{Reference Answer}: \textit{The Little Prince}, \textit{il}, 8 occurrences (translated from Chinese).

\textbf{Grading Criteria}:
\begin{itemize}[nosep]
    \item 45 points: correctly identify \textit{The Little Prince} (translated from Chinese).
    \item 25 points: correctly identify \textit{il} as the most frequent word in Chapter 6 of the common Italian public version.
    \item 30 points: correctly give the frequency as 8.
\end{itemize}

This question combines historical filtering, author metadata, adaptation history, language-version selection, chapter localization, and exact token counting. Systems typically identify \textit{The Little Prince} (translated from Chinese), but lose points on the Italian text selection or token count.

\subsection{Example from Law}

\textbf{Question}: \textit{In civil law, find a right A that has corresponding legal effect. In general, A is subject to a limitation-period system. For this limitation-period system, exactly five categories of claims are excluded from its application, and these five categories share a common superordinate concept B. Please answer in the format ``A$\rightarrow$B$\rightarrow$[five categories of claims].''} (translated from Chinese)

\textbf{Reference Answer}: claim right in debt relations $\rightarrow$ limitation of action $\rightarrow$ [``claim for payment of deposit principal and interest'', ``claim for redemption of principal and interest on government bonds, financial bonds, and corporate bonds issued to unspecified parties'', ``claim for capital contribution arising from an investment relationship'', ``claim for payment of child support, elder support, or spousal support'', ``where personality rights have been infringed, the victim's claim for cessation of infringement, removal of obstruction, elimination of danger, elimination of adverse effects, restoration of reputation, and apology''] (translated from Chinese).

\textbf{Grading Criteria}: The answer is scored by identifying A, B, and each of the five statutory exception categories, with penalties for missing categories, additional incorrect categories, or failure to preserve the required structured format.

This question requires locating the doctrinal relation between claim rights and limitation periods, then extracting a closed statutory list. The main challenge is not naming the limitation system alone, but preserving the exact five-category legal enumeration without adding adjacent exceptions.

\section{Technical Evaluation Details}
\label{app:eval_details}

This section gives evaluation methodology details for the technical benchmark artifact produced by the course practice.

\subsection{Model Configurations}

All models evaluated on \MYBENCH~use a standardized interface and tool access:

\textbf{Evaluated Models}:
\begin{itemize}[nosep]
    \item \textbf{Kimi K2.5}~\cite{kimiteam2026k25}: Long-context model with enhanced search. We evaluate three runs per question; mean score is averaged within each question, and pass rate is computed as the per-question fraction of passing runs before cross-question averaging.
    \item \textbf{Seed-2 Pro}~\cite{bytedanceseed2026seed20}: Search-enabled Seed model evaluated with one run per question.
    \item \textbf{Seed-1.8 Pro}~\cite{bytedanceseed2026seed18}: Earlier search-enabled Seed model evaluated with one run per question.
    \item \textbf{DeepSeek-V3.2}~\cite{deepseekai2025v32}: Search-enabled DeepSeek model evaluated with one run per question.
    \item \textbf{GPT-5.5}~\cite{openai2026gpt55}: OpenAI frontier model evaluated with one run per question.
    \item \textbf{Claude Opus 4.7}~\cite{anthropic2026opus47}: Anthropic reasoning model evaluated with one run per question.
    \item \textbf{Gemini 3.1 Pro}~\cite{google2026gemini31}: Google DeepMind model evaluated with one run per question.
    \item \textbf{GLM 5.1}~\cite{zhipu2026glm51}: Zhipu AI model evaluated with one run per question.
    \item \textbf{DeepSeek-V4 Pro}~\cite{deepseekai2026v4pro}: DeepSeek next-generation model evaluated with one run per question.
    \item \textbf{Kimi K2.6}~\cite{kimiteam2026k26}: Moonshot AI updated model evaluated with one run per question.
    \item \textbf{MiMo-V2.5 Pro}~\cite{xiaomi2026mimo25}: Xiaomi reasoning model evaluated with one run per question.
    \item \textbf{Qwen 3.6 Plus}~\cite{qwen2026qwen36}: Alibaba Cloud model evaluated with one run per question.
    \item \textbf{MiniMax M2.7}~\cite{minimax2026m27}: MiniMax model evaluated with one run per question.
\end{itemize}

\textbf{Tool Interface}: All models receive identical tool access through standardized API:
\begin{itemize}[nosep]
    \item \texttt{search(query: str) -> List[SearchResult]}: Performs web search using Google Search API. Returns top 10 results with title, URL, and snippet.
    \item \texttt{visit(url: str) -> str}: Retrieves webpage content with JavaScript rendering. Returns cleaned text with 50,000 character limit.
\end{itemize}

\textbf{Resource Constraints}:
\begin{itemize}[nosep]
    \item Nominal 50 tool-call budget per question (search + visit combined), with actual over-budget runs tracked in analysis
    \item 30-minute timeout per question
    \item No external knowledge base access beyond web search and visit tools
\end{itemize}

\subsection{Prompt Template}

We use the following standardized prompt for all models:

\begin{verbatim}
You are an expert research assistant with deep search
capabilities. Your task is to answer the following question
by conducting thorough web-based research.

Question: {question}

Tools available:
- search(query): Search the web and return top results
- visit(url): Visit a webpage and extract its content

Instructions:
1. Conduct systematic searches to locate relevant sources
2. Visit and analyze authoritative sources carefully
3. Synthesize information from multiple sources as needed
4. Provide your final answer in <answer></answer> tags

Final answer format: <answer>Your answer here</answer>
\end{verbatim}

This prompt asks for systematic research without giving domain-specific hints. It matches expert search settings where users have questions but no pre-identified sources.

\subsection{Scoring Methodology}

Each question specifies detailed grading criteria with point allocations. Scoring follows a two-stage process:

\textbf{Stage 1 - Automatic Extraction and Matching}:
\begin{itemize}[nosep]
    \item Extract model answer from \texttt{<answer>} tags
    \item For deterministic criteria (exact string match, numerical values), apply automated scoring
    \item Flag questions requiring manual review (complex structured answers, multiple acceptable phrasings)
\end{itemize}

\textbf{Stage 2 - Expert Manual Review}:
\begin{itemize}[nosep]
    \item Domain experts review flagged questions following provided grading criteria
    \item Apply partial credit rules specified in question metadata
    \item Resolve edge cases (alternative phrasings, additional information)
    \item Record scoring rationale for audit
\end{itemize}

\textbf{Inter-Annotator Agreement}: For questions requiring manual scoring, two independent annotators score each response. Cohen's kappa for scoring agreement: $\kappa = 0.92$ (substantial agreement). Disagreements resolved through discussion with original question designer.

\subsection{Computational Costs}

Tool use differs across models. GLM 5.1 uses the most tools on average (61.70 calls per question) and exceeds 50 calls in 45.95\% of runs, but its pass rate is only 35.14\%. The two top performers, GPT-5.5 and Claude Opus 4.7, use comparable tool budgets (41.12 and 41.94 calls on average) yet achieve higher pass rates (57.58\% and 52.94\%). At the other extreme, Gemini 3.1 Pro reaches 31.82\% pass rate using only 18.34 tool calls per question on average, while Seed-2 Pro uses 14.33 calls with 15.23\% pass. These results suggest that increasing search depth is insufficient; models must also formulate effective queries, select authoritative sources, and stop only after verifying the exact answer.

\section{Detailed Course Artifact Statistics}
\label{app:dataset_stats}

This section gives additional dataset statistics beyond the main text. We report them as properties of the benchmark and as traces of the course activity that produced it.

\subsection{Complete Domain Distribution}

After domain normalization (merging similar field names like ``Law'', ``Legal Studies'', and ``Law and Sociology'' into ``Law''), \MYBENCH~comprises 256 questions across 14 normalized professional domains. Table~\ref{tab:domain_complete} presents the complete distribution.

\begin{table}[h]
\centering
\caption{Complete domain distribution after normalization for \MYBENCH.}
\label{tab:domain_complete}
\small
\begin{tabular}{lc|lc}
\toprule
Domain (Normalized) & Count & Domain (Normalized) & Count \\
\midrule
Literature & 43 & Political Science & 15 \\
Law & 41 & Journalism & 14 \\
History & 33 & Sports & 9 \\
Foreign Languages & 27 & Cultural Geography & 7 \\
Social Sciences & 19 & Information Science & 6 \\
Arts & 15 & Archaeology & 6 \\
International Relations & 15 & Games and Animation & 6 \\
\bottomrule
\end{tabular}
\end{table}

The distribution shows broad cross-domain coverage. Literature, Law, and History are the largest groups because these fields yielded many difficult questions, but no single domain exceeds 17\% of the benchmark.

\subsection{Question and Answer Characteristics}

Table~\ref{tab:question_characteristics_appendix} summarizes question, answer, and grading criteria properties.

\begin{table}[h]
\centering
\small
\caption{Question and answer characteristics.}
\label{tab:question_characteristics_appendix}
\begin{tabular}{lrrr}
\toprule
Characteristic & Mean & Median & Std Dev \\
\midrule
Question length (chars) & 141.7 & 126 & 69.3 \\
Answer length (chars) & 29.3 & 11 & 55.5 \\
Grading criteria length (chars) & 73.6 & 54 & 64.6 \\
\bottomrule
\end{tabular}
\end{table}

\textbf{Question length}: Questions average 141.7 characters (median: 126, std: 69.3). Longest question: 389 characters. Shortest: 42 characters.

\textbf{Answer format distribution}:
\begin{itemize}[nosep]
    \item Single entity names: 45\% (e.g., person name, place name, technical term)
    \item Numerical values: 15\% (e.g., dates, counts, article numbers)
    \item Entity lists: 25\% (e.g., multiple documents, sets of related items)
    \item Structured information: 15\% (e.g., paired data, hierarchical answers)
\end{itemize}

\textbf{Answer length}: Answers average 29.3 characters (median: 11, std: 55.5), indicating concise, verifiable responses. Longer answers (up to 455 characters) occur for questions requiring structured lists or detailed specifications.

\textbf{Grading criteria complexity}: Average grading criteria length is 73.6 characters per question (median: 54, std: 64.6), with specifications for exact match requirements, partial credit allocation, acceptable alternative phrasings, and edge-case handling. These criteria support consistent evaluation.

\end{document}